\documentclass[sigconf]{acmart}

\AtBeginDocument{%
  \providecommand\BibTeX{{%
    \normalfont B\kern-0.5em{\scshape i\kern-0.25em b}\kern-0.8em\TeX}}}

\copyrightyear{2021}
\acmYear{2021}
\setcopyright{iw3c2w3}
\acmConference[WWW '21 Companion]{Companion Proceedings of the Web Conference 2021}{April 19--23, 2021}{Ljubljana, Slovenia}
\acmBooktitle{Companion Proceedings of the Web Conference 2021 (WWW '21 Companion), April 19--23, 2021, Ljubljana, Slovenia}
\acmPrice{}
\acmDOI{10.1145/3442442.3453455}
\acmISBN{978-1-4503-8313-4/21/04}

\usepackage{balance}

\begin{document}

\title[Developing Annotated Resources for Internal Displacement Monitoring]{Developing Annotated Resources for\\ Internal Displacement Monitoring}




\author[]{Fabio Poletto}  
\email{fabio.poletto@isi.it}
\affiliation[]{
\institution{ISI Foundation}
\city{Turin}
\country{Italy}
}

\author[]{Yunbai Zhang}  
\email{yz3386@columbia.edu}
\affiliation[]{
\institution{Columbia University} 
\city{Palisades}
\state{NY} \country{USA}
}

\author[]{Andr\'e Panisson}  
\email{andre.panisson@isi.it}
\orcid{0000-0002-3336-0374}
\affiliation[]{
\institution{ISI Foundation}
\city{Turin}
\country{Italy}
}

\author[]{Yelena Mejova}  
\email{yelena.mejova@isi.it}
\orcid{0000-0001-5560-4109}
\affiliation[]{
\institution{ISI Foundation}
\city{Turin}
\country{Italy}
}

\author[]{Daniela Paolotti}  
\email{daniela.paolotti@isi.it}
\orcid{0000-0003-1356-3470}
\affiliation[]{
\institution{ISI Foundation}
\city{Turin}
\country{Italy}
}

\author[]{Sylvain Ponserre}  
\email{sylvain.ponserre@idmc.ch}
\affiliation[]{
\institution{Internal Displacement Monitoring Centre}
\city{Geneva}
\country{Switzerland}
}


\renewcommand{\shortauthors}{Poletto et. al.}

\begin{abstract}
This paper describes in details the design and development of a novel annotation framework and of annotated resources for Internal Displacement, as the outcome of a collaboration with the Internal Displacement Monitoring Centre, aimed at improving the accuracy of their monitoring platform IDETECT. The schema includes multi-faceted description of the events, including cause, quantity of people displaced, location and date. Higher-order facets aimed at improving the information extraction, such as document relevance and type, are proposed. We also report a case study of machine learning application to the document classification tasks. Finally, we discuss the importance of standardized schema in dataset benchmark development and its impact on the development of reliable disaster monitoring infrastructure.
\end{abstract}



\maketitle


%
%
%



\section{Introduction}







The aim of this research is to contribute to the improvement of the technology for detection of a internal displacement (ID) in unstructured texts. We pursue this by developing a novel framework for annotating ID in online news articles and by applying it to five datasets covering three languages: English, Spanish, and French. 

Internal displacement (ID) is defined as the forced movement of people, who have to leave their home or place of habitual residence, within the country they live in. The major causes of ID are armed conflicts and disasters resulting from natural hazards, but other factors such as development plans, geophysical hazards or effects of climate change may as well displace people. Unlike migrants or refugees, Internally Displaced People (IDPs) do not cross any border and remain within their country. Depending on several factors, their condition may last indefinitely, and can get worse when institutions are not able, or willing, to provide assistance and protection, or when the place where they took shelter presents other risks. Although often neglected by media, ID is a massive phenomenon: in 2019 alone, 33.4 million new displacements have been recorded across 145 countries, of which about three fourths are caused by disasters resulting from natural hazards.  

\begin{figure}
	\centering
		\includegraphics[width=\linewidth]{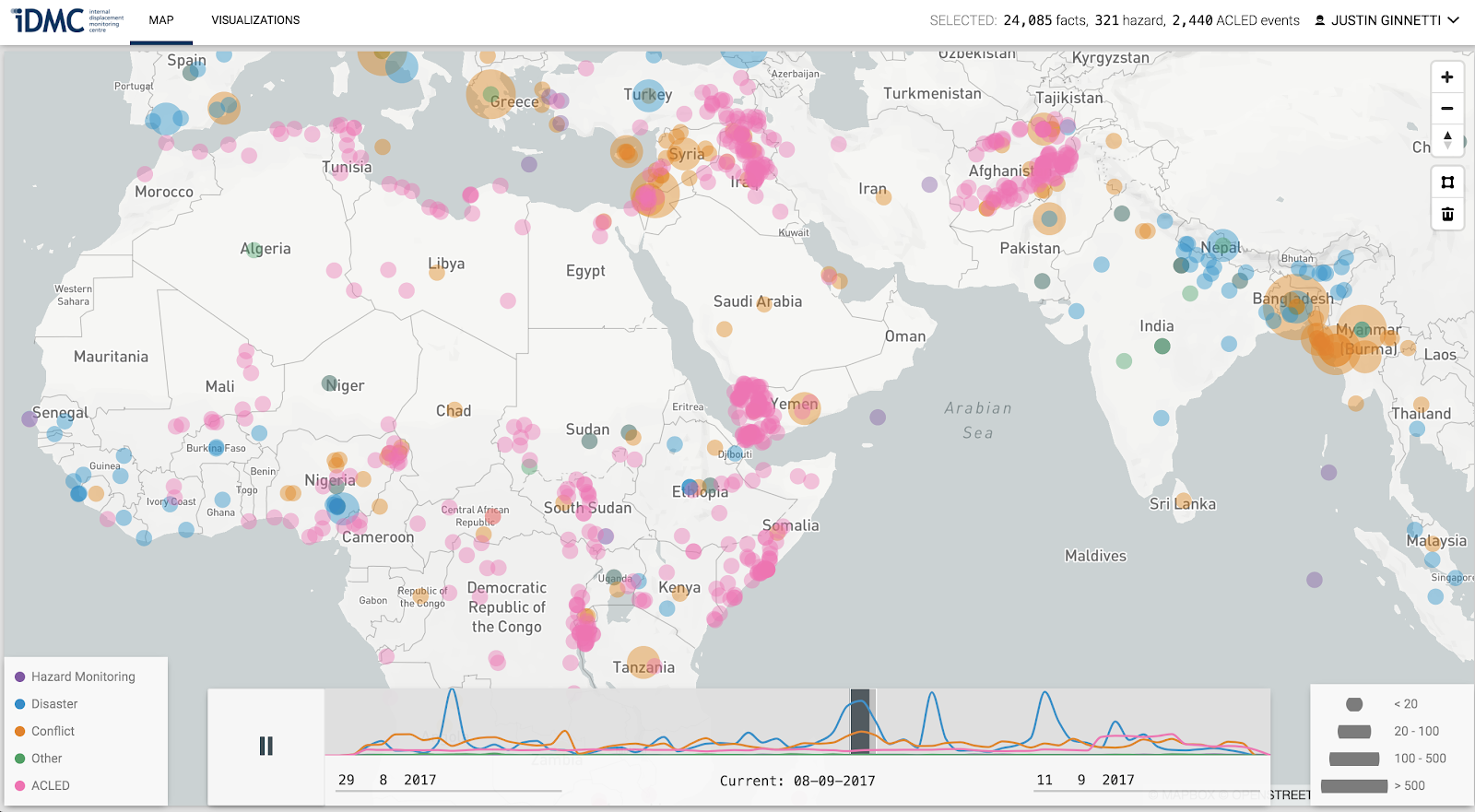}
	\caption{IDETECT interface showing detected instances of internal displacement and statistics over time.}
	\label{fig:idetectscreen}
\end{figure}

This work's rationale lies in a collaboration with Internal Displacement Monitoring Center (IDMC)\footnote{\url{https://www.internal-displacement.org/}}, an NGO dedicated to monitoring ID worldwide and to providing data, analyses and support to a broad range of international partners --- including the United Nations and the International Organization for Migration --- about how to tackle the phenomenon. Beside collecting data on ID from several official sources, IDMC makes use of a tool called IDETECT\footnote{\url{https://www.internal-displacement.org/monitoring-tools/monitoring-platform}}, an in-house platform that takes as input online news articles and reports related to displacement, processes the document with Natural Language Processing (NLP) and Machine Learning (ML) techniques, and outputs structured information about the ID event, including the type of event that caused it, where and when it took place, and how many people were involved. See Figure \ref{fig:idetectscreen} for an example of its interface.

Since this is a complex computational task and it is of utmost importance that the data and figures provided by IDMC are accurate, most of the platform's output is manually reviewed by a team of experts. This revision process is necessary, as the automated extraction of data is not always precise, with particular reference to dates, number of people affected, and locations.

We tackle this issue by designing a sophisticated framework for extracting all relevant information on ID from news articles, with a focus on solving the ambiguities and weak spots emerged from a careful analysis of the platform's output. We also used this framework to label five datasets: three of 200 articles each in English, Spanish and French, labeled by IDMC experts, one labeled with crowdsourcing on Amazon SageMaker (600 articles in English) and one labeled with crowdsourcing on Amazon Mechanical Turk (600 articles in English). The multilingual dataset was labeled by expert on a external platform called Kili\footnote{\url{https://kili-technology.com/}}, that allowed enhanced customization of the annotation task and an ongoing monitoring of the annotation quality for each user. 

We believe that our work has a twofold contribution. Firstly, the development of specific framework and resources for ID detection can contribute to the improvement of the platform's performance and, consequently, IDMC's concrete impact on monitoring and tackling this phenomenon. Secondly and more generically, releasing resources and an annotation framework for fine-grained event detection in unstructured texts can contribute to a field where there still is great space for improvements \cite{imran2015processing,soden2018informating}.

We conclude with an illustrative example of the application of machine learning to the problem of news article type classification (as \emph{news}, \emph{summary}, \emph{both} or none), and discuss the future directions and impact of the ongoing work.


\section{Related Work}

\subsection{Annotation Schemas}


Internal displacement monitoring tasks require automated information extraction from large amounts of unstructured texts.
Information extraction, on the other hand, requires annotated data for training and evaluation.
Named Entity Recognition (NER) and Disambiguation are two basic operations in this extraction process.
N$^3$~\cite{roder2014n3} is a collection of manually curated and annotated corpora stored in the NLP Interchange Format, focused on recognizing three main entity types: persons, organizations and places. Annotated corpus are mostly available only for English, and are often rare or limited for non-English languages in general. 
For the French language, an annotation schema for NER and relation extraction~\cite{jabbari2020french} is available for financial news.
It contains a custom NER annotation schema that contains basic entity types (Person, Organization, and Location) augmented with concepts and relations important to banking compliance.

%
The Event StoryLine Corpus (ESC)~\cite{caselli2017event} is a benchmark dataset for the temporal and causal relation detection that includes a task aimed at extracting and classifying events relevant for stories. The ESC adopts an event-centric annotation framework where the basic components of the annotation schema are events and temporal expressions, where an event is composed of an action, a time slot, a location and a participant that illustrates the ``who'' or ``what'' is involved in the action.

The purpose of such annotation schemas is to facilitate the creation of rich annotated datasets and remove uncertainty by making the process more objective. 
Nevertheless, annotating events is still difficult due to the complexity of the schemas and the ambiguity of the entities to be annotated.
Event extraction tasks are known to be more complex, given the lower measured performance in terms of F1-score compared to other types of entities such as people or locations, and the low inter-annotator agreement among experts when annotating events~\cite{inel2019validation}.
In many cases expert annotators are employed for this task, but non-expert annotated data might increase by a small margin the performance of extraction systems.
In our work, we similarly find the expert annotators to be vastly superior to the crowdsourced labelers.

\subsection{Disaster Monitoring Datasets}

Social media and microblogging platforms such as Twitter have been increasingly used in crisis situations and disasters caused by natural events, both for reporting, monitoring and coordinating response efforts. 
Twitter as a lifeline~\cite{imran2016twitter} is a human-annotated corpora with crisis-related Twitter messages collected during 19 different crises that took place between 2013 and 2015. 
The annotations are used to train classifiers to identify messages that are useful for humanitarian efforts. 
The datasets are also important for evaluating unsupervised approaches, such as one proposed by~\cite{tanev2017monitoring} that uses a natural language grammar learning approach to detect ``micro-events'' from news and social media with minimal human intervention, for monitoring disaster impact.

FacTA  (Factuality and Temporal Anchoring)~\cite{minard2015facta} is a task with associated evaluation dataset aimed to evaluate methods to (1) extract the ``factuality'' of events by associating event mentions to a set of attributes, namely: certainty, time, and polarity, and (2) to detect events for which it is possible to identify temporal relations between events and temporal expressions.

Other sources of data, such as the technology use and mobile phone data, can help in tracking mobility changes in real-time.
For instance,~\cite{lu2012predictability} use a large mobile phone dataset to track the impact of the 2010 Haiti earthquake, and show that the destinations of people who left the capital was correlated highly with the locations in which people had significant social bonds.
Similarly,~\cite{wilson2016rapid} use call records to to track the mass movements immediately after the 2015 Nepal Earthquake from the Kathmandu valley to the surrounding areas.

Further, for topics especially difficult to track, communities have turned to citizen science and crowdsourcing. For instance, the Gun Violence Database~\cite{pavlick2016gun} is built and updated through a continuously running crowdsourced annotation pipeline with daily crawls of local newspapers and television websites from across the US describing incidents of gun violence.

Using these many sources of data, some of which may be crowdsourced, and others processed automatically, it may be possible to alleviate the lack of information about the disasters and ID events in the non-Western countries, which has been well documented in the news coverage \cite{franks2006carma}. 
Although the annotation schema presented in this work concerns news, it may be easily adapted to other sources of data, in order to aid automated information extraction from multiple data sources. 





  
\subsection{GDELT}

Introduced at a greater length in the Data Collection section, Global Data on Events, Location, and Tone or GDELT is a massive collection of worldwide broadcast news that has been extensively used in research\footnote{\url{https://www.gdeltproject.org/}}~\cite{leetaru2013gdelt}.
For instance, it has been used to show the biases in world news coverage of disasters, with a focus on the parts of the world which are politically less stable~\cite{kwak2014first}.
It has been compared to other datasets for event monitoring, such as the ICEWS early warning system, finding that the fine-grained geolocation may be useful for disaggregation of data~\cite{ward2013comparing}.
Finally, it has been used to study attention dynamics around particular topics, including climate change, and compared to social media, pointing to differences in triggers, actions, and news values between the information streams~\cite{olteanu2015comparing}.




  








\section{Data Collection}


Since we proposed to improve a system that extracts knowledge primarily from long, unstructured texts written by humans --- generally press articles or reports --- we decided to develop and test our methodology on similar data. Therefore, we draw from the same source as IDMC monitoring system: GDELT~\cite{leetaru2013gdelt}. GDELT is a massive database which collects the vast majority of worldwide broadcast news, in more than a hundred languages, and leverages NLP techniques to classify and analyse them. Among other things, the classification performed by GDELT assigns labels to each article according to its detected themes, along with other text entities such as counts, people names, organization names, locations, emotions, relevant imagery, video, and embedded social media posts. IDMC makes use of this annotation to collect any document having at least one label that may be somehow related to displacement. For example, any news mentioning displacement, evacuation, relief camps, people or buildings affected by disasters is collected. The goal of this methodology is to make sure not to miss important documents, ensuring high recall: further classification will then sort documents based on their relevance.
Table~\ref{table:theme-occurrences} reports the list of themes extracted from articles collected in 2019, with their number of occurrences in the data. Note that the most common themes extracted by \textsc{GDELT} include the keywords \textit{refugees}, \textit{evacuation} and \textit{displaced}, and other themes related to 
Economic Policy Uncertainty (EPU) categories~\cite{baker2016measuring}
and to the CrisisLex repository of crisis-related social media data and tools~\cite{olteanu2014crisislex}.



\begin{table}[h!]
\begin{center}
\begin{tabular}{l@{}r@{}r@{}}
\toprule
Theme &  \#refs &  \#docs\\
\midrule
Refugees &      676.442 &      421.274 \\
Evacuation &      470.961 &      360.441 \\
CrisisLex T09: Displaced and evacuated ppl. 
&      411.020 &      318.040 \\
EPU Cats Migration Fear Migration &      249.788 &      188.187 \\
Displaced &      215.606 &      181.381 \\
CrisisLex C05: Camp and shelter &      165.542 &      117.145 \\
\midrule
Total &2.189.359&\hspace{0.2cm}1.027.922\\
\bottomrule
\end{tabular}
\end{center}
\caption{\label{table:theme-occurrences}
List of themes extracted from online news articles collected in 2019. \#refs shows the total number of references, while \#docs shows the number of articles containing the theme. An article might contain references to many themes, and more than one reference to the same theme.}
\end{table}

Despite the abundance of information extracted by GDELT, the only information that IDETECT stores is the URL to the original article. The content is then retrieved and analyzed independently, using a NLP pipeline that first classifies each document by relevance and event category, extracts its publication date and then, for relevant articles, extracts facts, locations, dates and number of affected people. 
The output from IDETECT is combined and triangulated with data from outside sources in order to achieve estimates as accurate and inclusive as possible of the number and conditions of displaced people everywhere in the world almost in real time. The result of all this process is then used to inform the Global Internal Displacement Database\footnote{\url{https://www.internal-displacement.org/database}} (GIDD), a publicly available platform that provides up-to-date statistics and visualization on all the ID events in the world. 
  
These data are the starting point of our work. Upon manual inspection and validation of a sample, we noticed some flaws in the output data when compared to the full text of the article. In particular, the correct classification by relevance and the extraction of locations and number of individuals affected appeared to be often inaccurate. This observation, and the need to improve the accuracy of IDETECT, led us to rethink the information extraction framework on which the system is based and to work on a novel, detailed framework that could reduce the chance of error on difficult categories. 

Furthermore, although GDELT automatically translates most of non-English articles into English, IDETECT only retrieves and processes articles that are originally published in English. We wanted to widen its range by providing labeled data in other languages. For this purpose, we collected articles in French and Spanish following the same procedure used for English articles.   


We present a collection of five distinct datasets, differing from each other in either size, language and scope. Three datasets have been created for expert annotation on the platform Kili: they contain 200 articles each in English, French and Spanish respectively (these will be henceforth referred to as Kili-EN, Kili-FR and Kili-ES). Two more dataset have been created for crowdsourcing annotation on Amazon SageMaker and on Amazon Mechanical Turk (henceforth, Amazon-SM and Amazon-MT). They contain 600 English articles with three annotations per article, for a total of 1.800 annotations collected in each platform.
All the documents have been retrieved using the same methodology, which is the one used by IDETECT: URLs were collected from GDELT by filtering relevant topics, then their content was scraped and stored together with the publication date. We used the same list of keywords\footnote{Keywords are as follows: \emph{displaced, evacuated, forced, flee, homeless, relief camp, sheltered, relocated, stranded, stuck, accommodated, destroyed, damaged, swept, collapsed, flooded, washed, inundated, evacuate, families, person, people, individuals, locals, villagers, residents, occupants, citizens, households, home, house, hut, dwelling, building, Rainstorm, hurricane, tornado, rain, storm, earthquake, refugee camp, refugee center, refugee, asylum seeker, crossed, arrived, entered, evicted, eviction, sacked}} for filtering results from GDELT. From an initial collection of several hundred thousands, for each one of the five datasets we filtered a random sample covering the whole year 2019. 


\section{Annotation Framework}

Two separate annotation schemes were designed for each of the platform used to collect annotations, due to their different structure and to the different targets that were expected to use each platform. These two schemes are based on the same general framework, and represent respectively a complete, detailed version and a simplified version of it. They will henceforth be referred to as Expert scheme and Crowdsourcing scheme.

The Expert scheme was designed first, leveraging the close collaboration with the team of researchers from IDMC and making several adjustments until the scheme has been deemed complete and accurate enough by the whole team. Since we were able to count on a team of monitoring experts who are familiar with concepts related to ID and with analyzing and identifying relevant information from newspaper articles, we could develop a complex and multi-layer framework with the aim of maximizing the accuracy of the information extracted. The structure and boundaries of this framework also reflect the range of resources offered by the tool we used to design it. The final scheme consists of nine labeling tasks, including versions of  Classification, Named Entity Recognition and Named Entity Relation.

For the Crowdsourcing scheme we adapted and simplified the previous scheme, omitting all the Named Entity-related tasks and breaking down the remainders to the simplest possible form. There are two reasons behind this somewhat drastic choice: the platforms on which this scheme was meant to be used are much less flexible than Kili and do not allow for the complexity we needed; and the labelers did not receive any training on the task nor on the subject except for the instructions we provided, therefore we could not expect the same understanding and accuracy.

In both cases, the annotators were provided with detailed guidelines, including general instructions on how to annotate successfully, a careful description of each task and a list of the labels available for each task. 

\subsubsection*{Expert scheme}

In this framework there are nine labeling tasks of three different types. Two of them are Classification tasks and concern the whole document, six are Named Entity Recognition tasks concern single words or phrases, while one is a Named Entity Relation task and operates on other labels. Furthermore, two tasks included a nested Transcription task allowing the labeler to type more detailed information in a fixed format.

Only the two Classification tasks were required --- i.e., for each document, users had to select a label for these tasks in order to be able to submit their annotation --- while all the others were optional and could be skipped. Nonetheless, in our guidelines we recommended to use all the labels for which relevant content could be found in the document. Labels could overlap partially or totally, so that the same word could bear multiple labels expressing different information, if necessary.  

\begin{figure*}
	\centering
		\includegraphics[width=0.8\linewidth]{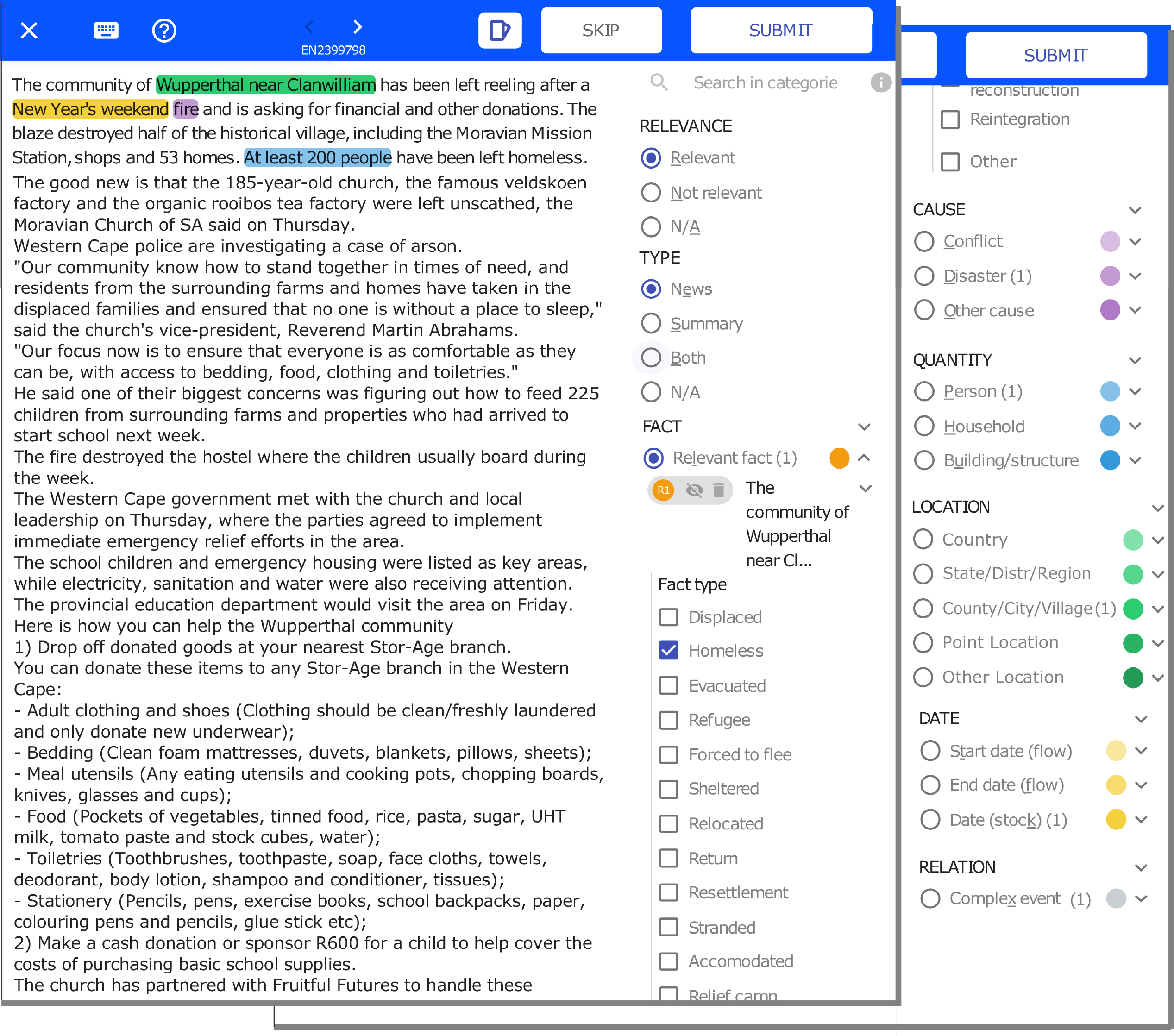}
	\caption{Annotation interface with the text to be annotated (left) and the categories and labels (right). Due to the long list of categories available, we show the interface split in two screens. The text to be annotated is the same in the two screens.}
	\label{fig:kiliscreen}
\end{figure*}

Below, the nine tasks are briefly presented, indicating for each one its type, category and labels available. 

\begin{itemize}

    \item \textbf{Relevance} -- whether the article contains any relevant information about ID or IDPs. Labels: \textit{Relevant, Not Relevant, N/A}.
    If a document was labeled as \textit{Not Relevant} or \textit{N/A}, the user was allowed to skip all the optional tasks and submit his/her annotations with only the two required labels.
    
    \item \textbf{Type} -- the type of article on the basis of its content, distinguishing: reports on recent events (up to four days back); comments or updates on events extended over time or happened more than four days back; articles displaying both features. Labels: \textit{News, Summary, Both, N/A}. These first two tasks were required and concerned the whole document.
    
    \item \textbf{Relevant fact} -- The core fact reported by the article, where a fact is relevant if it gives information about ID or IDPs. For each fact, one or more Fact Type label could be selected from a list of 18 labels, specifying the kind of displacement.
    
    \item \textbf{Cause} -- what kind of situation or event triggered the displacement, if reported. Labels: \textit{Conflict, Disaster, Other cause}.
    
    \item \textbf{Quantity} -- the number of entities involved in the displacement, if reported. Labels: \textit{Person, Household, Building/Structure}. A nested Transcription task here allowed the labelers to type in the exact number and, if any, qualifier used to determine the count such as ``more than", ``about" or ``at least".
    
    
    
    \item \textbf{Location} -- the location where the displacement happened, i.e., the place people are moving or have moved away from, if reported. Labels: \textit{Country, State/District/Region, County/City/-Village/Town/Hamlet, Point Location, Other Location}. 
    
    \item \textbf{Location Destination} -- the location where the displacement is directed to, i.e., the place people are moving or have moved towards. The labels are identical to Location Origin.
    
    \item \textbf{Date} -- when the displacement started or the time span it covered. Labels: \textit{Start Date (flow), End Date (flow), Date (stock)}. A nested Transcription task here allowed the labelers to type in the exact date reported in the article, according to the format "YYYYMMDD".
    
    \item \textbf{Complex event} -- this task, unlike the others, required to establish a relation between pairs of Named Entity labels applied before. Linking together the elements spread across the text would allow to reconstruct the whole relevant event.
    The relation was constrained inasmuch as the first element was necessarily a label "Relevant fact", while the second element could be any other label except "Relevant fact". This would lead to the creation of one-to-many relations, where all the existing information about an ID fact (why, how many, where, when) where linked to the fact itself, regardless of their distance inside the document.

\end{itemize}

There was no limit to the number of labels to be used in a single document, but a few observations need to be done. None of the task allowed the choice of multiple labels for the same item, except for Fact Type. Thus, a phrase labeled as Relevant Fact could have ``evacuated" and ``forced to flee" as Fact Type, but a phrase labeled as cause could only be either ``conflict", ``disaster" or ``other": in order to express two causes, two different labels had be to placed.
However, some constraints were put on the ``Relation" label between Named Entities. Firstly, they could only be one-to-many relations between a ``Relevant fact" label was linked to any other label referring to the same event, except other ``Relevant fact" labels. Secondly, every label had to be linked to a ``Relevant fact" label, and there could be no stand-alone labels lying outside a relation with a ``Relevant fact"\footnote{This is, in fact, rather a constraint on the general labeling process, as experts had to be careful throughout the job not to create labels that would have then remained unrelated.}. In other words, information about causes, counts, locations and dates of an ID event can not exist without an ID event itself. A summary of all the tasks with their characteristics is provided in Table \ref{tab:expertscheme}.

\begin{table}[ht]
\centering
\begin{tabular}{lcccr}
\hline
\textbf{Task} & \textbf{Type} & \textbf{Requi-} & \textbf{Multiple} & \textbf{Labels} \\ 
&& \textbf{red} & \textbf{labels} & \\
\hline 
Relevance & CL & YES & NO & 3 \\ 
Type & CL & YES & NO & 4 \\ 
Fact & NER & NO & NO & 1 \\
    * \textit{Fact type} & \textit{MC} & \textit{NO} & \textit{YES} & 18 \\
Cause & NER & NO & NO & 3 \\ 
Quantity & NER & NO & NO & 3 \\
    * \textit{Count} & \textit{TR} & \textit{NO} & \textit{NO} & -- \\
    * \textit{Qualifier} & \textit{TR} & \textit{NO} & \textit{NO} & -- \\
Location & NER & NO & NO & 5 \\  
Date & NER & NO & NO & 3 \\
    *  \textit{YYYYMMDD date} & \textit{TR} & \textit{NO} & \textit{NO} \\
Complex event & NERel & NO & YES & -- \\ 

\hline
\end{tabular}
\caption{\label{tab:expertscheme}Summary of all the annotation tasks in the Expert scheme. Rows in italics introduced by an asterisk indicate nested task. Shortenings in the Type column: CL = classification, NER = named entity recognition, MC = multiple choice, TR = transcription, NERel = named entity relation.}
\end{table}

\subsubsection{Crowdsourcing scheme}


For the crowdsourced annotation we broke the Expert scheme down to its simplest form. Setting up Named Entity Recognition tasks on the two Amazon platforms was either not possible or not effective for our purpose, and the same applied to the Named Entity Relation task. Therefore we only kept the two Classification tasks, and merged them into one for further simplicity. This resulted in a single-task annotation process structured as follows:

\begin{itemize}
    \item \textbf{Classification} -- whether the article contains any relevant information about ID or ID persons and, if so, whether it reports recent or less recent events, or both. Labels: \textit{Relevant -- News, Relevant -- Summary, Relevant -- Both, Not Relevant, N/A}.
\end{itemize}

Multiple choice was not allowed for this task.

\subsection{Annotation process} 

\subsubsection{Expert annotation on Kili}

While developing our annotation framework we realized that it was assuming such size and complexity that it would not fit into a simple spreadsheet or into a traditional crowdsourcing format, but it would require a dedicated tool with great flexibility. We therefore turned to Kili, a platform for annotation that allowed us to shape the project to our needs, to have full control on all the stages of the annotation, to monitor the progress and the quality of the process and to set up inter-annotator agreement. Control at all stages was possible through web interface and through API.

A total of six expert annotators participated in the projects, five of which from IDMC team and one from ISI team. All of them contributed to the annotation of the Kili-EN dataset, while only four labeled the Kili-FR and Kili-ES datasets, being native or fluent speakers of those languages.  
A subset of the Kili-EN dataset was used for consensus: 20\% of the data (40 articles) were set to be annotated by three annotators,
in order to collect information on the agreement trends without burdening the team with excessive workload. No consensus was set on the other two dataset, since annotators were the same and we could assume similar behaviours across the different projects. 

The annotators were provided with extended instructions --- which they actually contributed refining with their feedback --- including a detailed description of each task and instructions on how to perform the annotation and deal with a variety of special cases and issues. The guidelines were easily accessible at any time throughout the annotation. 

The annotation was performed through the intuitive Kili web interface. Annotators were presented one article at a time, with the list of tasks and labels always visible on a side panel. It was possible to display the publication date of each article, which was often necessary for the Date transcription task. Once all the articles were labeled, annotators had a thorough feedback session discussing the main issues and doubts. They then took a second round of annotation, reviewing their labels with the aim to ensure consistency throughout the dataset.

\subsubsection{Crowdsourced annotation on Amazon}


\begin{figure*}
	\centering
		\includegraphics[width=0.8\linewidth]{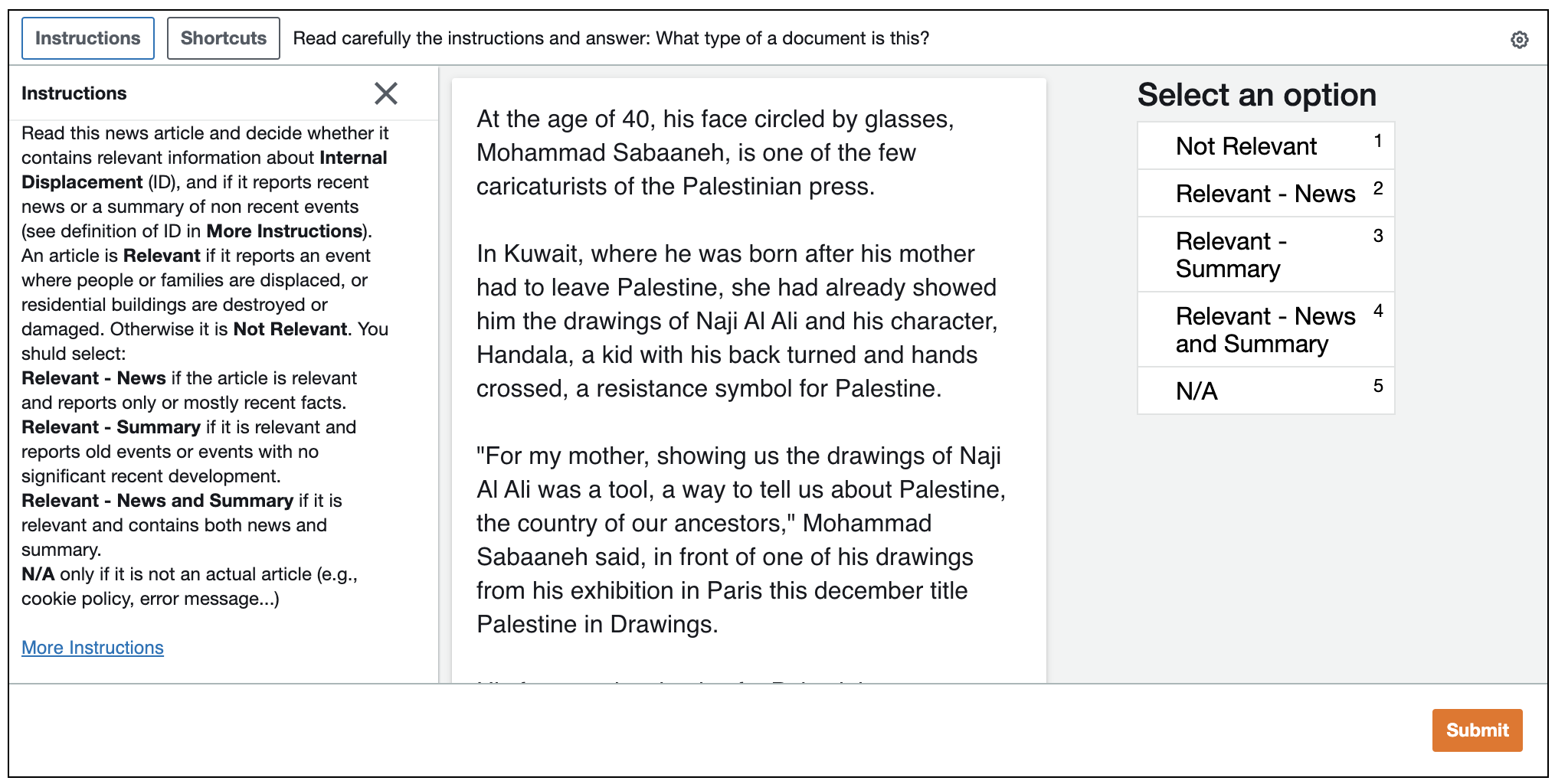}
	\caption{Crowdsourced annotation task on Amazon Mechanical Turk.}
	\label{fig:awsscreen}
\end{figure*}

Two separate crowdsourcing tasks were set up: one on Amazon SageMaker and one on Amazon Mechanical Turk (AMT). Both platforms use external, untrained workforce to label data. Such tools provide quick annotations but less flexibility in modeling the tasks, and we expected the overall quality to be lower. 
We set a filter that only allowed reliable users to label data (e.g. with high approval rate), and for the AMT dataset we required each article to be labeled by three judges. 

We provided a much shorter version of the guidelines, that only presented the most important instructions in a straightforward way. Figure~\ref{fig:awsscreen} shows a screenshot with one of the annotation assignments on AMT.

Data annotation process on SageMaker works as follows: after configuring the annotation interface with layout and instructions, the requester uploads the data to be annotated and awaits until all tasks are completed.
The process on AMT is more flexible: the requester can monitor the annotation process, approve or reject assignments submitted by workers while the annotation is still in progress, and block workers from receiving more assignments if they do not pass quality assessments. When an assignment is rejected, it returns to the pool to be reassigned.
This larger flexibility in principle should allow for a better selection of workers and enforce higher annotation quality.
This fine-tuned monitoring of the annotation process, to the present date, is not yet available on SageMaker.


\section{Quality assessment}

Since we started our project with the aim of building quality resources for Internal Displacement, once the annotation stage was completed we set up to measure the robustness and reliability of what we had collected. Kili provides an overall percentage agreement among labelers for the data with multiple annotations, that in the case of the Kili-EN dataset was 67\%. Nonetheless, we wanted to find a more robust and explainable metric that takes into account the peculiarities of each task. 

The job was all but easy, as our data included a variety of different annotation tasks: the number of labels available varied from task to task; some applied to the document level, others combined text selection and labeling, while others still required some typing; some included nested tasks, and one required to apply links instead of labels. As for the Expert dataset, we computed agreement on the articles labeled by three annotators (40) and, among them, only on those that were labeled as \textit{Relevant} by at least two annotators (19). Since our guidelines prescribe not to annotate all the other categories if an article is labeled as \textit{Not Relevant}, we were only interested in those for which we could compare at least two complete annotations. 

Eventually we opted for Krippendorff's $\alpha$, deeming it the most appropriate metric. Due to technical reasons we could not apply it to all the tasks: we discarded the three transcription tasks (two nested in the Quantity task and one nested in the Date task), and the Named Entity Relation task (Complex Event). Among the remaining tasks, we had two cases: pure Classification tasks and Named Entity Recognition tasks. 

As for the formers (Relevance and Type), we compared the labels chosen by each labeler, as it is referred to the whole document. 
For the latters (Cause, Quantity, Location Origin, Location Destination, Date), in order to achieve a more accurate measure, we compared separately the label chosen and the position of the selected Named Entity (NE) within the document, for each task. 

We used a nominal metric of the type \textit{a = b} for comparing labels, except for the task Type, for which we created a tailored metric due the particular nature of the labels. In fact, we consider the label \textit{Both} to match \textit{News} and \textit{Summary}.

An issue that emerged in the comparison of NEs is that some articles, especially the longer ones, might repeat the same information more than once in the text, often with almost identical words. After discussing with the team of expert from IDMC we realized that sometimes two annotators had selected the same information but at different points in the document. This would result in a complete disagreement, although they actually identified and labeled the same concept. Therefore, beside computing the overlap between position indexes of two labels for a given task, we recomputed it adding a similarity metric between the text. This allowed us to state that two annotators are in agreement for a given task, if they assigned the same NE label to two different strings with different positions in the text but very similar content.

Finally, we merged together the agreement based on labels and the agreement based on overlap and similarity as follows: first we checked if there was any overlap between two selected entities, then, if there was no overlap, we checked for text similarity; lastly, we compared labels only between entities having either overlap or similarity, discarding those having none of the two. This two-steps process avoided directly comparing labels that were assigned to completely different text portions, which would make little sense when evaluating agreement on Named Entities annotation and would most likely result in a higher disagreement score.

\begin{table*}[ht]
\centering
\begin{tabular}{lccccl}
\hline
\textbf{Task} & \textbf{$\alpha$ labels} & \textbf{$\alpha$ overlap} & \textbf{$\alpha$ overlap + sim.} & \textbf{$\alpha$ merged} & \textbf{Metric} \\
\hline 
Relevance & 0.72 & - & - & - & nominal \\
Type & 0.58 & - & - & - & tailored \\
Fact & 0.44 & 0.58 & 0.58 & 0.50 & nominal \\    
Cause & 0.81 & 0.33 & 0.66 & 0.72 & nominal \\
Quantity & 0.20 & 0.54 & 0.60 & 0.52 & nominal \\
Location Origin & -0.02 & 0.39 & 0.55 & 0.30 & nominal \\
Date & 0.42 & 0.08 & 0.29 & 0.33 & nominal \\
\hline
\end{tabular}
\caption{Annotator agreement measures over different tasks in the English-language dataset. \label{tab:alphas}}
\end{table*}
Table \ref{tab:alphas} shows the $\alpha$ score for each task. The task Location Destination does not appear as no label for that task was assigned in our sample; for the task Type, being it a Classification task, only the metric on labels could be computed. 

In future work, we will perform similar quality assessments of annotations in the other languages, and compare the performance to that of English. Note that the same assessors were working on several languages, which leads us to speculate that the performance in the other languages will be similar. 

Furthermore, from preliminary analysis, we find the annotator agreement (and quality) to be vastly poorer for the crowdsourced annotation (with scores closer to 0). We will continue to experiment with such platforms in order to improve label quality, but suggest to the readers to find dedicated labelers, resources permitting.




\section{Application of Machine Learning}

As an illustrative example, we present preliminary results of machine learning experiments concerning the Document Type task, wherein we identify whether the document is \emph{news}, \emph{summary}, \emph{both}, or none of these. After excluding documents that were labeled as non-relevant, and resolving label conflicts using the majority rule, 95 documents remain for analysis. The class proportions are dominated by one class: \emph{news} with 62 documents, \emph{summary} at 24, and \emph{both} at 9. 

We use TF-IDF weighting of n-grams of n between 1 and 5, and only those which occur in at least 5 documents. 
In practice we are interested not in the three-class problem, but to detect articles that have any relevant news. Thus we consider a two-class problem distinguishing between whether a document is (\emph{news} or \emph{both}) versus a \emph{summary}.
Using 10-fold cross-validation, we test several classifiers, listed in Table \ref{tab:kind_classifiers}. We find that the performance comparable for all the classifiers, with AUC of around 0.620 on a test set, except SVM, which consistently under-performs. 

\begin{table}[ht]
\centering
\begin{tabular}{lcc}
\hline
& \textbf{Validation} & \textbf{Test} \\
\textbf{Classifier} & \textbf{mean Au-ROC} & \textbf{AUC} \\
\hline 

Multinomial Naive Bayes & 0.651 & 0.621 \\
XGBoost                 & 0.631 & 0.605 \\
Random Forest           & 0.648 & 0.617 \\
Logistic Regression     & 0.641 & 0.618 \\
Support Vector Machine  & 0.643 & 0.623 \\
\hline
\end{tabular}
\caption{Performance of Document Kind classification task. Validation performance during tuning is computed via 10-fold cross-validation, and test performance is computed once on a test set. Average over 50 train/test splits. \label{tab:kind_classifiers}}
\end{table}

Further, we apply the same algorithms to the document Relevance classification task, where a document may be \emph{relevant} (we have 91 such documents in the English corpus), or \emph{not-relevant} (102 documents). Table \ref{tab:relevance_classifiers} shows the performance of the classifiers, indicating that the performance is much better than document type. 

\begin{table}[ht]
\centering
\begin{tabular}{lcc}
\hline
& \textbf{Validation} & \textbf{Test} \\
\textbf{Classifier} & \textbf{mean Au-ROC} & \textbf{AUC} \\
\hline 

Multinomial Naive Bayes & 0.827 & 0.823 \\
XGBoost                 & 0.756 & 0.774 \\
Random Forest           & 0.813 & 0.825 \\
Logistic Regression     & 0.829 & 0.829 \\
Support Vector Machine  & 0.828 & 0.827 \\
\hline
\end{tabular}
\caption{Performance of Document Relevance classification task. Measures computed as in Table \ref{tab:kind_classifiers}.  \label{tab:relevance_classifiers}}
\end{table}



\section{Discussion \& Conclusions}

Leveraging digital news to gain real time insight on phenomena that affect individuals and society has become a widespread approach in many areas, from epidemiology of infectious diseases to disasters and conflict preparedness. Especially in the latter context, collecting and analyzing a wide array of information to create up-to-date-reports on population displacement has become an invaluable effort for many humanitarian agencies. By mining huge news data sets, such as GDELT, it is possible to capture significantly more real-time data on the effect of disasters and conflict on population displacement. 
In this work, we have presented an approach aimed at improving an existing platform, IDETECT, which relies on Natural Language Processing routines collecting information from digital news to provide alerts of displacement events. The goal of the work was to provide a more effective annotation framework aimed at identifying domain-specific types of events with high accuracy in time and space as well as in the number of individuals affected by the event. The schema presented in this work has been developed in collaboration with the experts at the IDMC, and thus presents a valuable insight into the information necessary for the high-quality and detailed reporting on ID. 
A wide number of distinctions among the ID persons was developed, to capture the particular circumstances of the event. 
It is possible that in the future, using automated techniques, even more distinctions can be extracted from the texts, especially if the schema is to be applied to non-news text sources (such as first-hand accounts via social media).

The next step in this project will involve the evaluation of the existing IDETECT NLP pipeline using this annotated dataset, in order to quantify the quality of extracted information. 
The same information can then be used to test different NLP/ML approaches to information extraction, especially for the tasks shown to be least accurate in the existing system.
The multi-lingual nature of the dataset can also be utilized in assessing the possibility of applying automatic machine translation in order to apply non-native NLP tools to texts. 

The process that led to the creation of the high-quality annotated data set discussed in this work has shown some important limitations. First, annotators were required to perform very sophisticated and complex tasks. Even though the annotators were experts in the specific domain of internal displacement, it required considerable effort to come up with adequate guide lines that would describe the required tasks as clearly as possible without ambiguity. Despite this effort, the required tasks would present a significant degree of arbitrariness in the way information could be selected by the annotators. This has led to a lower than expected degree of alignment across annotators which has presented some challenges in the choice for the evaluation metrics. The complexity of the tasks also poses issues about the replication of the effort in the future, in case the need for updated annotated data arises (e.g. to avoid concept drift issues).
Finally, it remains to be assessed whether it is worthy to repeat this effort in other languages different from English and build a language-specific machine learning framework or instead translate everything into English and apply the already existing one.

In conclusion, mining digital news through systems such as IDETECT can significantly enhance the capability of humanitarian agencies such as IDMC to identify more reported incidents of internal displacement and create real-time actionable reports. On the other hand, IDETECT was not conceived to replace traditional data collection and analysis, but to complement it and improve it.
Continuous collaboration among national authorities, UN and other international organisations with research institutions and specialised media can play a key role in the monitoring process by combining traditional sources as well as novel digital ones. 

\section{Data Availability}

The annotated and documented data can be found at \url{https://github.com/ISIFoundation/Internal-Displacement-Monitoring}. The link also includes the code for assessing the quality of the annotations and the evaluation of the machine learning models presented in this paper.

\section*{Acknowledgments}

\balance

The authors would like to thank Álvaro Sardiza Miranda,
Cl\'ementine Andr\'e,
Elisabeth du Parc Locmaria,
Maria Teresa Miranda Espinosa and
Marta Lindström,
from the Internal Displacement Monitoring Centre,
for their valuable contributions as expert annotators.
The authors would like to thank
Justin Ginnetti and
Alex de Sherbinin
for supporting the collaboration among the institutions.
The authors acknowledge support from the Lagrange Project
of the Institute for Scientific Interchange Foundation (ISI
Foundation) funded by Fondazione Cassa di Risparmio di
Torino (Fondazione CRT).

\bibliographystyle{ACM-Reference-Format}
\bibliography{displacement}

\end{document}